\documentclass{article}
\usepackage{spconf,amsmath,graphicx}
\usepackage{amsfonts}

\usepackage{algorithmic}
\usepackage{algorithm}
\usepackage{xcolor}

\usepackage{subcaption}
\usepackage{pifont}

\newtheorem{theorem}{Theorem}

\title{Deep Adversarial Defense Against Multilevel-$\ell_p$ Attacks}
\name{Ren Wang$^{\star}$ \qquad Yuxuan Li$^{\star}$ \qquad Alfred Hero$^{\dagger}$\thanks{Alfred Hero was supported in part by the US National Science Foundation under Grant CCF-2246213  and the US Army Research Office under Grant W911NF-23-1-0343. Ren Wang was supported by the US National Science Foundation under Grant 2246157 and the ORAU Ralph E. Powe Junior Faculty Enhancement Award.}}
\address{$^{\star}$ Dept. ECE, Illinois Institute of Technology $^{\dagger}$ Dept. EECS, University of Michigan}

\begin{document}
%
\maketitle

\begin{abstract}
Deep learning models have shown considerable vulnerability to adversarial attacks, particularly as attacker strategies become more sophisticated. While traditional adversarial training (AT) techniques offer some resilience, they often focus on defending against a single type of attack, e.g., the $\ell_\infty$-norm attack, which can fail for other types. This paper introduces a computationally efficient multilevel $\ell_p$ defense, called the Efficient Robust Mode Connectivity (EMRC) method,  which aims to enhance a deep learning model's resilience against multiple $\ell_p$-norm attacks. Similar to analytical continuation approaches used in continuous optimization, the method blends two $p$-specific adversarially optimal models, the $\ell_1$- and $\ell_\infty$-norm AT solutions, to provide good adversarial robustness for a range of $p$. We present experiments demonstrating that our approach performs better on various attacks as compared to AT-$\ell_\infty$, E-AT, and MSD, for datasets/architectures including: CIFAR-10, CIFAR-100 / PreResNet110, WideResNet, ViT-Base.
\end{abstract}


\section{Introduction}
\label{sec:intro}

Deep learning models have revolutionized numerous fields, offering innovative solutions to complex problems \cite{li2023physics,jouaiti2022dysfluency}. However, their vulnerability to adversarial attacks remains a significant concern \cite{goodfellow2014explaining,wang2022ask,sun2022adversarial}, undermining their practical utility and reliability. Specifically, these models are sensitive to slight, yet strategic, perturbations in their input data, which can mislead them into making incorrect predictions. While several methods aim to defend against such adversarial manipulations, most focus on enhancing the model's resilience against attacks based on a single type of perturbation metric, often measured by a $\ell_p$ norm \cite{madry2018towards,shafahi2019adversarial,wang2020fast,Wong2020Fast} for specific $p\in [1,\infty]$. This focus creates a defensive blind spot, leaving models vulnerable to other types of adversarial perturbations. On the other hand, recent studies that aim to achieve universal robustness across multiple $\ell_p$ norms either suffer from high computational costs or do not entirely solve the universal robustness problem: maintaining robustness against different types of perturbations concurrently \cite{croce2019provable,stutz2020confidence,maini2020adversarial}.

This paper addresses the shortcomings of current methods by proposing universally robust models capable of countering diverse types of $\ell_p$-norm adversarial attacks. Previous studies have identified the mode connectivity property, which suggests that a path of high accuracy and low loss exists between two well-trained models in the parameter space \cite{freeman2016topology,garipov2018loss,wang2023exploring}. Building on this concept and the theoretical evidence that affine classifiers can withstand multiple types of $\ell_p$ attacks if they are already resistant to $\ell_1$ and $\ell_\infty$ perturbations \cite{croce2019provable}, our work presents a novel approach: Efficient Robust Mode Connectivity (ERMC) combined with Model Ensemble which waeves $\ell_1$ and $\ell_\infty$ robustness into the fabric of mode connectivity to derive a new training methodology. This amalgamation enables the identification of parameter paths that remain highly resistant to both $\ell_1$ and $\ell_\infty$ perturbations, and therefore, multiple types of $\ell_p$ norm perturbations. We introduce an optimized fine-tuning technique with reduced computational complexity. Lastly, we employ a model ensemble strategy to select and aggregate models from this robust path, further improving robustness. Specifically, the algorithm works as follows. We first train one endpoint model optimized for $\ell_\infty$-norm adversarial training then retrain the model to be optimal relative to the $\ell_1$-norm. Using these two endpoint models, and leveraging the mode connectivity property of deep neural networks (DNN), we identify a low-loss, high-robustness path connecting these endpoints. Finally, we deploy ensemble model aggregation to select models along this path that exhibit collective robustness against all types of $\ell_p$-norm attacks, $1\leq p\leq \infty$. 


\noindent\textbf{Contributions.} We summarize our contributions below.
\begin{enumerate}
    \item We improve upon traditional mode connectivity approaches to the design of DNN by integrating adversarial robustness, thereby uncovering a path that links an $\ell_\infty$ and an $\ell_1$ adversarially trained model. This path demonstrates high resistance to other $\ell_p$-norm attacks for $p \in [1,\infty]$.
    \item We propose an Efficient Robust Mode Connectivity (ERMC) method, supplemented with model ensemble aggregation, that results in an efficient adversarial training algorithm with enhanced robustness.
    \item Numerical experiments demonstrate that the proposed ERMC with model ensemble has superior performance in robustness against various $\ell_p$ attack modalities when compared to baseline approaches.
\end{enumerate}

\noindent The rest of this article is structured as follows: Section~\ref{sec:relatedwork} introduces related research on single-attack adversarial strategies and countermeasures, as well as defenses against a variety of $\ell_p$ norm perturbations. The subsequent Section~\ref{sec: multi} on multilevel $\ell_p$-defense delves into the specifics of adversarial training and the optimization of robustness against multilevel $\ell_p$ perturbations. It lays the theoretical groundwork for our approach and addresses the question of achieving concurrent high robustness against both $\ell_1$ and $\ell_\infty$ perturbations. Section~\ref{sec:method} presents our novel ERMC approach in detail, describing how it incorporates robustness into mode connectivity and the ensemble model strategy used to boost robustness. Section~\ref{sec:exp} reports on the datasets, model architectures, evaluation methods, and the comprehensive experimental results, showcasing the effectiveness of ERMC compared to established methods. The Conclusion Section~\ref{sec:conclusion} summarizes our findings and contributions.

\section{Background and Related Work}
\label{sec:relatedwork}

\subsection{Adversarial Attacks And Defenses}
Recent studies have revealed that conventional machine learning models are susceptible to adversarially modified datasets. For a model $\boldsymbol \theta$ an adversary can target each feature $\mathbf x \in \mathbb{R}^d$ in a database $\mathcal D$ of feature-label pairs $\mathcal D=\{{\mathbf x},y\}$, by solving the 
following {\em attacker's optimization} problem:
\begin{align}\label{eq: adv_atk}
    \displaystyle \arg\max_{\mathbf x'} {\mathcal L(\boldsymbol \theta; {\bf x}^\prime, y)}, ~~ s.t.~~ d_p({\bf x}^\prime, {\bf x}) \leq \epsilon_p.
\end{align}
Here $\mathcal L$ represents the training loss, e.g., the cross-entropy loss. $\epsilon_p$ is the attack-strength parameter of the type-$p$ attacker, and $d_p$ is a distance metric of type-p over the model parameter space. As in many other studies, we restrict attention to the case that $d_p$ is the $\ell_p$ norm with $p \in [1, \infty]$. The solution to \eqref{eq: adv_atk} is commonly known as the $\ell_p$ adversarial attack \cite{madry2018towards}. This problem is often iteratively solved using the fast gradient sign method \cite{goodfellow2014explaining} or projected gradient descent (PGD) \cite{madry2018towards}, which computes the gradient $\nabla_{\mathbf x'}\mathcal L(\mathbf \theta; {\mathbf x'},y)$ combined with a projection that constrains the perturbation ${\bf x}^\prime - {\bf x}$ to the $\ell_p$-ball of radius $\epsilon_p$. The projection for the $\ell_p$ adversarial attack is denoted by $P_{\boldsymbol \epsilon_p}$. These methods may result in suboptimal solutions to \eqref{eq: adv_atk} due to incorrect hyper-parameter tuning and gradient masking. To address these issues, methods such as the Auto Attack (AA)  \cite{croce2020reliable} and Multi Steepest Descent (MSD) \cite{maini2020adversarial} were introduced. Adversarial attacks can operate in a black-box manner, meaning the attacker does not have access to the model's parameters \cite{andriushchenko2020square,chen2017zoo}. However, this paper focuses only on scenarios where the attacker is aware of the model's parameters. To counter the attackers strategy \eqref{eq: adv_atk}, adversarial training (AT) methods are effective defense mechanisms  \cite{madry2018towards,shafahi2019adversarial,wang2020fast,Wong2020Fast}. However, these methods often focus on a single type of $\ell_p$ disturbance, leading to decreased robustness against different types of perturbations~\cite{tramer2019adversarial}.

\subsection{Robustness Towards Multiple $\ell_p$ Norm Perturbations}
In ~\cite{stutz2020confidence} the authors propose training on $\ell_\infty$-generated adversarial examples while selectively discarding inputs having low confidence scores, showing empirically that this results in a degree of robustness to $\ell_p$-attacks for $p=0,1,2,\infty$. The authors of \cite{tramer2019adversarial} propose calculating the worst case attack by either picking the attack type that leads to the maximum loss or averaging the loss across all attack types. The Multi Steepest Descent (MSD) Defense \cite{maini2020adversarial} integrates multiple perturbation schemes to yield a more comprehensive $\ell_p$ robustness. The work in \cite{croce2019provable} offers a theoretically guaranteed defense mechanism but it only applies to affine classifiers. The Extreme Norm Adversarial Training (E-AT) method \cite{croce2022adversarial} employs a form of fine-tuning to practically implement the pathway from \cite{croce2019provable} and to reduce AT computational load. 
In contrast, in this paper we exploit the mode connectivity property of deep neural networks \cite{freeman2016topology,garipov2018loss,wang2023exploring} to define the ERMC method that improves on the performance reported in \cite{croce2022adversarial}.

\section{Multilevel $\ell_p$-defense}\label{sec: multi}

\noindent\textbf{Adversarial training (AT).} Complementing the attacker's optimization \eqref{eq: adv_atk}, the defender aims to solve the {\em defender's optimization} problem:   
\begin{align}
\begin{array}{ll}
    \displaystyle \min_{\boldsymbol \theta} \mathbb E_{(\mathbf x, y) \in \mathcal D} \left[ \displaystyle \max_{\mathbf x^\prime: d_p({\bf x}^\prime, {\bf x}) \leq  \epsilon_p} 
 \mathcal L( \boldsymbol \theta; \mathbf x^\prime,  y ) \right],
 \end{array}
 \label{eq: at}
\end{align}
using training data from $\mathcal D$ to empirically estimate the statistical expectation in \eqref{eq: at}, resulting in a solution we call Adversarial Training (AT)-$\ell_p$. The main issue addressed in this section is that the solution AT-$\ell_p$ for a given $p$ does not ensure robustness to other values of $p$ in $[1,\infty]$.  Furthermore, while in principle one could compute a dense set of solutions $\{$AT-$\ell_p\}_{p\in [1,\infty]}$, it is not clear how such solutions could be computed and combined in a computationally tractable manner to provide robustness over a range of $p$ \cite{tramer2019adversarial}. 

\noindent\textbf{Optimizing robustness against multilevel $\ell_p$ perturbations.} As argued in ~\cite{croce2019provable}, affine and piecewise affine classifiers (like CNN with ReLU) can resist multiple $\ell_p$ norm attacks if they are already robust to $\ell_1$ and $\ell_\infty$ perturbations. Specifically, Theorem 3.1 in ~\cite{croce2019provable} states that the convex hull of the union ball of the $\ell_1$ and $\ell_\infty$ provides satisfactory robustness to $\ell_p$ perturbations, $1 \le p\le \infty$:
\begin{theorem}
    \cite{croce2019provable} Suppose that the classifier is piecewise affine. Let $C$ be the convex hull of the union ball of the $\ell_1$ and $\ell_\infty$. If $d\ge 2$ and $\epsilon_1 \in (\epsilon_\infty, d\epsilon_\infty)$, then
    \begin{equation}
  \min_{\mathbb{R}^d\backslash C} \|\mathbf x^\prime - \mathbf x\|_p = \frac{\boldsymbol \epsilon_1}{(\boldsymbol \epsilon_1/\boldsymbol \epsilon_\infty - \beta + \beta^q)^{1/q}}
    \end{equation}
where $\beta=\frac{\boldsymbol \epsilon_1}{\boldsymbol \epsilon_\infty} - \lfloor \frac{\boldsymbol \epsilon_1}{\boldsymbol \epsilon_\infty}\rfloor$ and $\frac{1}{p}+\frac{1}{q}=1$.
\end{theorem}

The salient question arising is: how can one concurrently achieve high robustness against both $\ell_1$ and $\ell_\infty$ perturbations? To address this question a cutting-edge study, E-AT \cite{croce2022adversarial}, proposed using a method called fine-tuning to efficiently update the model from AT-$\ell_\infty$ to AT-$\ell_1$, asserting that this results in robustness to a range of $\ell_p$ disturbances. Yet, two notable issues persist: \noindent\ding{182} while the fine-tuned model may exhibit robustness against $\ell_1$-norm attacks, it may have lost some robustness against the original $\ell_\infty$-norm attack; and \noindent\ding{183} Achieving both high $\ell_\infty$ robustness and high $\ell_1$ robustness is inherently challenging for a single model, given its limited capacity. To address these dual challenges, this paper introduces a mode-connectivity-based approach that simultaneously identifies a large number of models having both high $\ell_\infty$ robustness and high $\ell_1$ robustness. This results in a larger union ball, thereby enhancing the model's resilience against a broader range of perturbations.

\section{Proposed Methods}
\label{sec:method}
We aim to improve the joint robustness to both $\ell_\infty$ and $\ell_1$ perturbations by leveraging two adversarially trained models. 

\subsection{Incorporating robustness into mode connectivity}

For neural networks, mode connectivity is the property that pairs of local minima (modes) discovered by gradient-based optimization techniques are connected through simple paths over which the model's loss does not change appreciably \cite{freeman2016topology,garipov2018loss}. In \cite{garipov2018loss} mode connectivity is established for a wide range of DNNs and training datasets. The path between a pair of modes $\boldsymbol \theta_1, \boldsymbol \theta_2 $ is constructed over the parameter space of the neural network by minimizing the averaged loss function, $\mathcal L$, over all possible simple paths. The path is represented as $\phi_{\boldsymbol \theta}=\{\phi_{\boldsymbol \theta}(t), t \in [0,1]\}$, where $\boldsymbol \theta$ is a free parameter (control point),  which satisfies the endpoint conditions $\phi_{\boldsymbol \theta}(0)=\boldsymbol \theta_1$ and $\phi_{\boldsymbol \theta}(1)=\boldsymbol \theta_2$.  Specifically, to find a desired low-loss path between the modes $\boldsymbol \theta_1$ and $\boldsymbol \theta_2$, one minimizes the following statistical expectation
\begin{equation}
\min_{\boldsymbol \theta} \mathbb E_{t \sim U(0,1)}  \displaystyle \mathbb E_{({\bf x},y) \sim \mathcal D}  \mathcal L( \phi_{\boldsymbol \theta}(t); ({\bf x},y)),
\label{eq: mc}
\end{equation}
where $U(0,1)$ represents the uniform distribution over the interval $[0,1]$.
The curve $\phi_{\boldsymbol \theta}$ is fixed as a Quadratic Bezier Curve (QBC) \cite{farouki2012bernstein} across this paper:
\begin{equation}
\begin{aligned}
   &\phi_{\boldsymbol \theta}(t) = (1-t)^2 \boldsymbol \theta_1 + 2t(1-t)\boldsymbol \theta + t^2 \boldsymbol \theta_2.
\end{aligned}
\label{eq: curves_QBC}
\end{equation}

The main assumption behind this paper is that the notion of mode connectivity can be extended to adversarial loss functions associated with different $\ell_p$-types, resulting in paths that maintain a high level of robustness against both $\ell_\infty$ and $\ell_1$ attacks, in addition to improving robustness to other $\ell_p$ attacks.  The proposed extension consists of two additional steps: \textbf{(Step 1)} The endpoint parameters $\boldsymbol \theta_1$ and $\boldsymbol \theta_2$ are trained via AT-$\ell_\infty$ and AT-$\ell_1$; \textbf{(Step 2)} We solve the following modification of \eqref{eq: mc} to preserve adversarial robustness for $p\in\{1,\infty\}$:  
\begin{equation}
\min_{\boldsymbol \theta} \mathbb E_{t \sim U(0,1)}  \displaystyle \mathbb E_{({\bf x},y) \sim \mathcal D } \sum_{p \in \{1,\infty\}} \max_{d_p({\bf x}^\prime, {\bf x}) \leq  \epsilon_p} \mathcal L( \phi_{\boldsymbol \theta}(t); ({\bf x}^\prime,y)),
\label{eq: mc_adv}
\end{equation}
where $\phi_{\boldsymbol \theta}(0)$ and $\phi_{\boldsymbol \theta}(1)$ are the two AT models, AT-$\ell_\infty$ and AT-$\ell_1$, respectively. In the inner optimization loop $d_p$ corresponds to the $\ell_\infty$ and $\ell_1$ distances for $p=0$ and $p=1$. 
We use a Multi Steepest Descent (MSD) technique to solve the maximization in the inner loop that encompasses both $\ell_\infty$ and $\ell_1$ perturbations within each step of PGD, similarly to \cite{maini2020adversarial}. In each epoch, for every data batch, we randomly choose a value for $t$. The subsequent training closely resembles Adversarial Training (AT), with the key difference being that we pick the worst-case perturbation from two types of perturbations in each inner loop iteration. Consequently, the algorithmic complexity remains similar to that of standard AT.

\begin{algorithm}[h]
\caption{Efficient Robust Mode Connectivity}
\label{alg: ERMC}
\begin{algorithmic}[1]
\REQUIRE A model $\phi_{\boldsymbol \theta}(0)$ trained with AT-$\ell_\infty$; initial model $\boldsymbol \theta^0$; the corresponding projections $P_{\boldsymbol{\delta}_1}$ and $P_{\boldsymbol{\delta}_\infty}$; training set $\mathcal D$; iteration number $J$; batch size $B$; initial perturbation $\boldsymbol{\delta}^{(0)}=\mathbf 0$.
\STATE{Create a copy of $\phi_{\boldsymbol \theta}(0)$ and retrain it with AT-$\ell_1$ for 10 epochs to obtain a model $\phi_{\boldsymbol \theta}(1)$.}
\STATE{$\boldsymbol \theta = \boldsymbol \theta^0$}
\FOR{each data batch $\mathcal D_b \in \mathcal D$ in each epoch $e \in E$}
\STATE{Uniformly select $t \sim U(0,1)$}
\FOR{$\forall \bf x \in \mathcal D_b$}
\FOR{$j = 1, \cdots, J$}
\STATE{\hspace{-1mm}$\boldsymbol{\delta}_1^{(j)} \leftarrow P_{\boldsymbol \epsilon_1}\big(\boldsymbol{\delta}^{(j-1)}-{\nabla_{\boldsymbol{\delta}}\mathcal L( \phi_{\boldsymbol \theta}(t); {\bf{x}}+{\boldsymbol{\delta}}^{(j-1)},y)}\big)$}
\STATE{\hspace{-1mm}$\boldsymbol{\delta}_\infty^{(j)} \leftarrow P_{\boldsymbol \epsilon_\infty}\big(\boldsymbol{\delta}^{(j-1)}-{\nabla_{\boldsymbol{\delta}}\mathcal L( \phi_{\boldsymbol \theta}(t); {\bf{x}}+{\boldsymbol{\delta}}^{(j-1)},y)}\big)$}
\ENDFOR
\STATE{$\boldsymbol \delta^{(j)} \leftarrow \arg\max_{\boldsymbol \delta_i^{(j)}, i \in \{1,\infty\}} {\mathcal L( \phi_{\boldsymbol \theta}(t); {\bf{x}} + \boldsymbol \delta_i^{(j)}, y)}$}
\ENDFOR
\STATE{$\boldsymbol \theta \leftarrow \boldsymbol \theta - \nabla_{\boldsymbol \theta} \sum_{\bf x \in \mathcal D_b} \mathcal L(\phi(t; \boldsymbol \theta); {\bf x} + \boldsymbol \delta^{(j)}, y)$}
\ENDFOR
\RETURN $\boldsymbol \theta$, $\phi_{\boldsymbol \theta}(t), \forall t \in [0,1]$
\end{algorithmic}
\end{algorithm}

We conclude this sub-section by noting that the concept of expansion of the set of high adversarially robust models beyond two models AT-$\ell_1$ and AT-$\ell_\infty$ is similar to the concept of analytic continuation in complex analysis, more specifically the converse analytic continuation method called blending, which seeks to extend two analytic functions defined over disjoint domains to a single $C^{\infty}$ function over a path connecting the domains \cite{trefethen2023numerical}.

\subsection{ERMC with model ensemble}

\begin{figure*}[t]

\begin{minipage}[b]{.24\linewidth}
  \centering
  \centerline{\includegraphics[width=5.0cm]{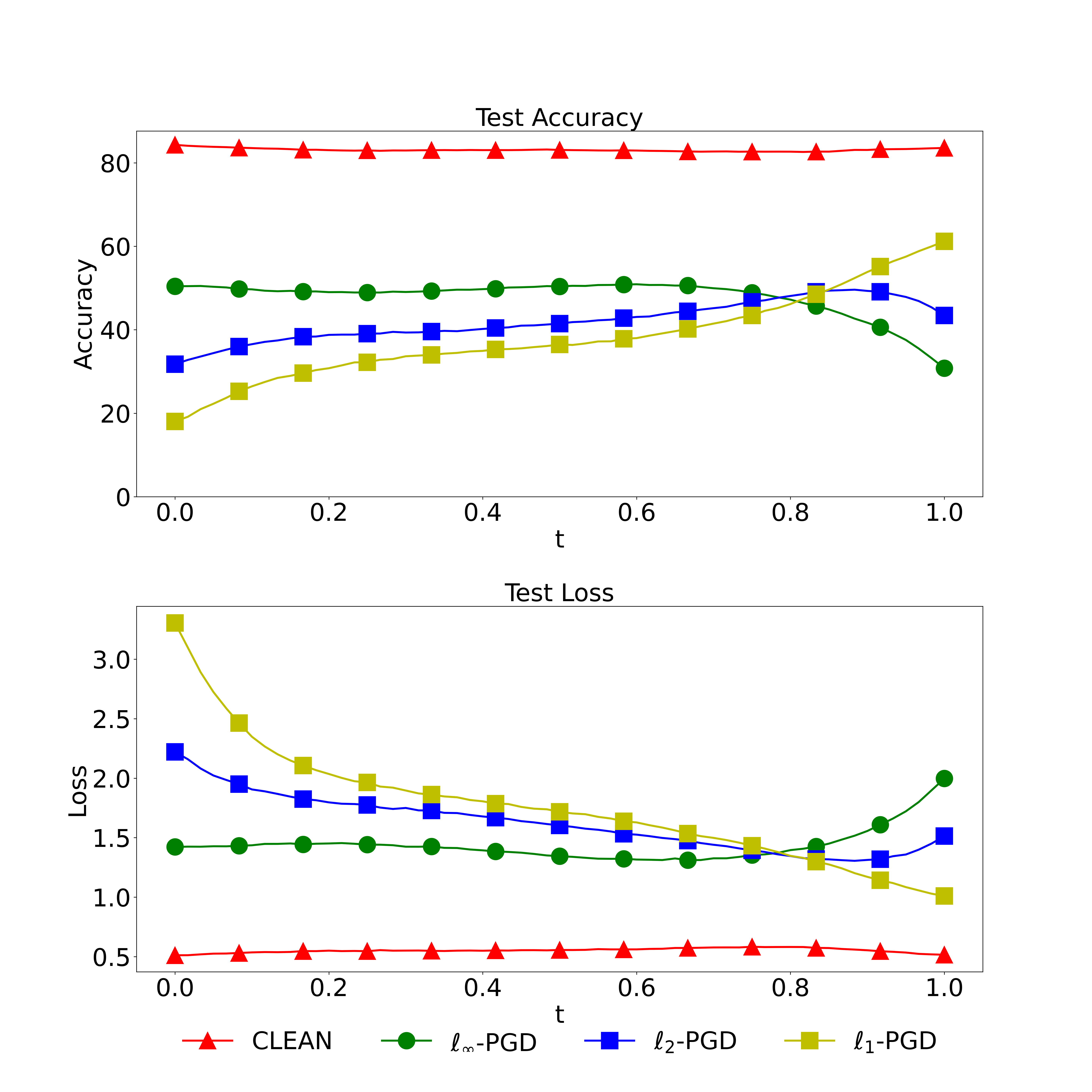}}
  \centerline{(a) CIFAR-10}\medskip
\end{minipage}
\begin{minipage}[b]{.24\linewidth}
  \centering
  \centerline{\includegraphics[width=5.0cm]{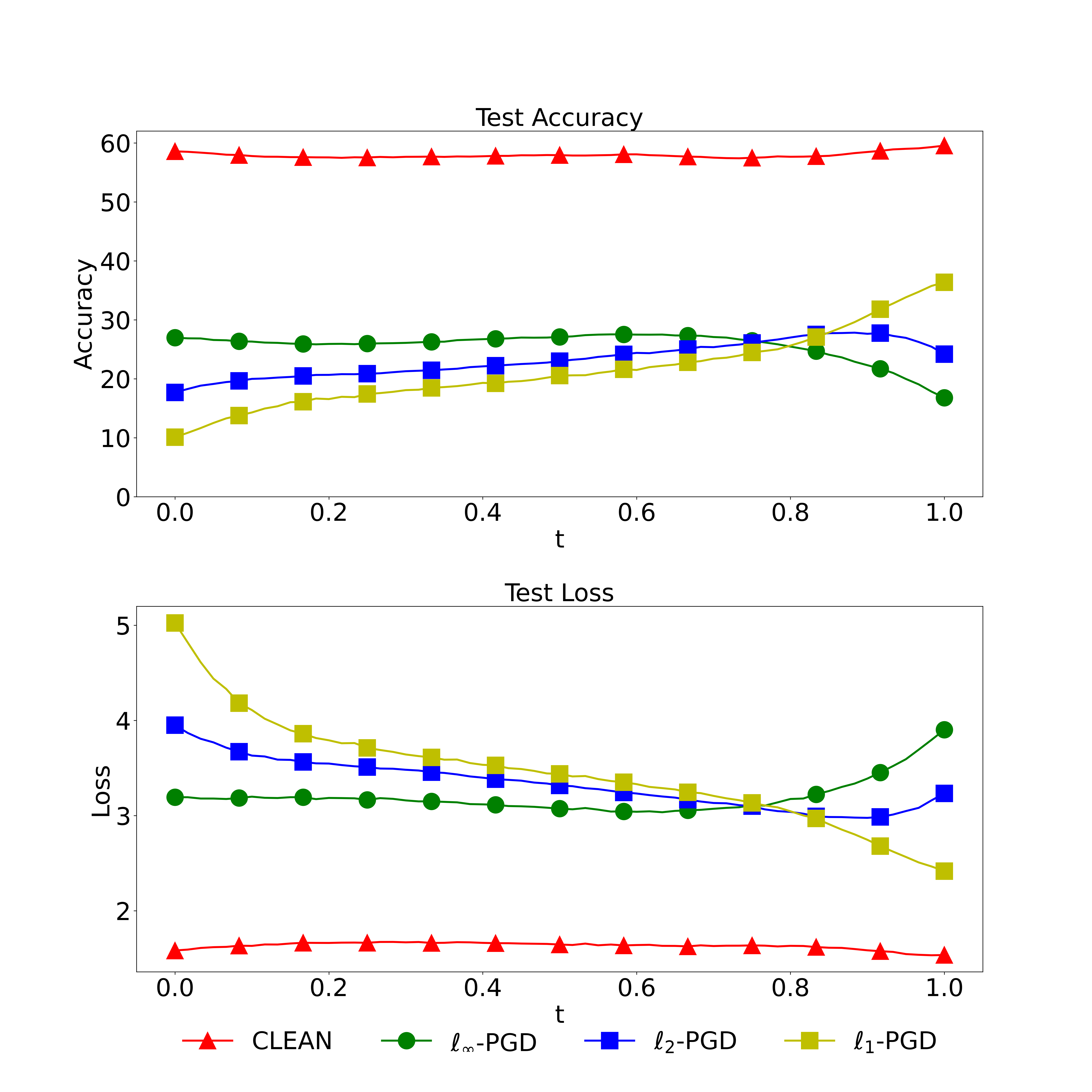}}
  \centerline{(b) CIFAR-100}\medskip
\end{minipage}
\begin{minipage}[b]{.24\linewidth}
  \centering
  \centerline{\includegraphics[width=5.0cm]{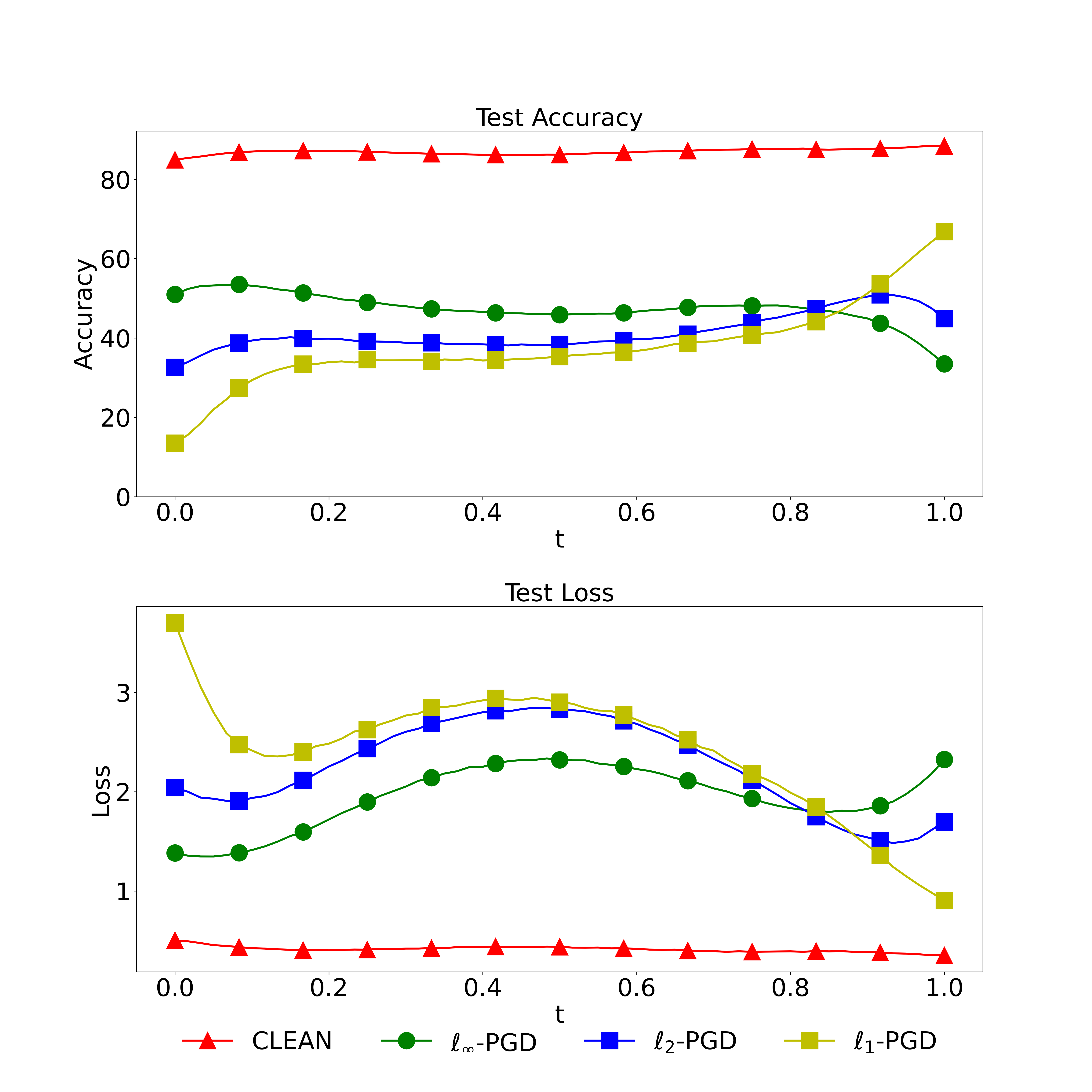}}
  \centerline{(c) WideResNet-28-10}\medskip
\end{minipage}
\hfill
\begin{minipage}[b]{0.24\linewidth}
  \centering
  \centerline{\includegraphics[width=5.0cm]{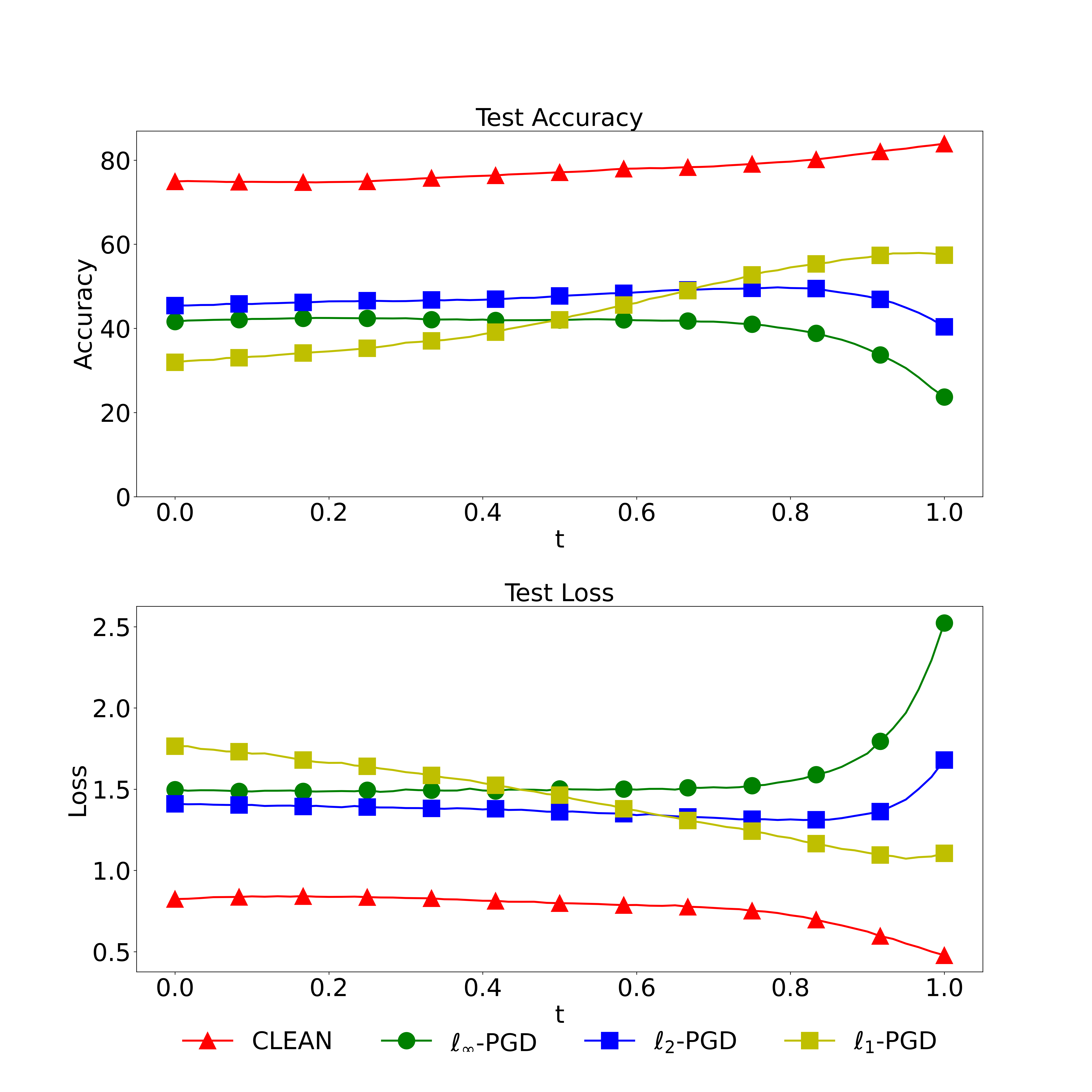}}
  \centerline{(d) ViT-base}\medskip
\end{minipage}
\caption{\small ERMC can find paths with high robustness against $\ell_\infty/\ell_2/\ell_1$ attacks by connecting a $\ell_\infty$ model and a $\ell_1$ model. The effectiveness of ERMC is validated on different datasets and model architectures. Upper panels: the accuracy of the clean test and the robust accuracies under $\ell_\infty/\ell_2/\ell_1$-PGD attacks. Lower panels: the associated loss values of clean test data and perturbed test data. (a) and (b): results obtained from the CIFAR-10 and CIFAR-100 datasets, using the PreResNet110 model architecture. (c) and (d): results from the CIFAR-10 dataset, utilizing the WideResNet-28-10 and ViT-base model architectures.}
\label{fig: adv_self_rmc2}
\end{figure*}

We reduce the computation burden of solving two independent AT-$\ell_p$ problems, for $p=\infty$ and $p=1$, by introducing a more efficient approach: the efficient robust model connectivity algorithm. 
%
In ERMC, initially a model with high robustness to either $\ell_\infty$ or $\ell_1$ perturbation is trained, {after which a copy is created and retrained for another few epochs under the other perturbation type using the same training set. Similarly to its use in E-AT  \cite{croce2022adversarial}, the fine-tuning step provides more efficient computation of the AT-$\ell_\infty$ and AT-$\ell_1$ adversarial models in EMRC. } In the experiments described below, the number of fine-tuning epochs is set to $10$ yielding a computationally less burdensome determination of the second endpoint model, while retaining the first one, facilitating the identification of a high-robustness path as provided by \eqref{eq: mc_adv}. 
The full algorithm of ERMC is presented in Algorithm~\ref{alg: ERMC}. The second endpoint $\phi_{\boldsymbol \theta}(1)$ is trained from the first endpoint $\phi_{\boldsymbol \theta}(0)$ using a different perturbation type. In each epoch, we sample a $t$ uniformly from the uniform distribution. Then, in each iteration of generating perturbations, we consider two types: $\ell_\infty$ and $\ell_1$. Subsequently, we select the perturbations that cause the highest losses and use them to update the model parameters. The number of iterations, denoted by $J$, is set at 10 for our experiments.

We have observed in experiments that certain regions along the path contain models that exhibit high levels of robustness for both types of perturbations. The optimal model along the path can be identified by assessing the trajectory with lower robust accuracy under $\ell_\infty$ and $\ell_1$ attacks, and then selecting the point that performs best in this worst-case scenario. This single-model approach offers the advantage of circumventing the limitations inherent to E-AT \cite{croce2022adversarial} while capitalizing on robustness against both types of perturbations. However, given the existence of many models along the path that exhibit high degrees of robustness to $\ell_\infty$ and $\ell_1$ attacks, it's natural to consider a model ensemble strategy to further bolster performance. This leads to a model that is collectively more robust to both $\ell_\infty$ and $\ell_1$ perturbations. The ensemble selection proceeds as follows. We find a segment $[a,b]$ along the path $\phi_{\boldsymbol \theta}$ satisfying the criterion: each point on the segment has robust accuracies surpassing two prefixed {\it model selection thresholds} $\alpha_\infty, \alpha_1$ under $\ell_\infty$ and $\ell_1$ attacks, respectively. 
We then choose $n > 1$ models at path locations given by $t = a + \frac{b - a}{n - 1}i$, where $i$ ranges from $0$ to $n - 1$. If multiple non-continuous intervals meet the above criterion, the $n$ points can be distributed among them proportionately to their respective lengths. {We denote ERMC with $n$ selected models as ERMC-$n$ and average the outputs of these $n$ models' final layers to form our class probability prediction.}


\section{Experiments}
\label{sec:exp}

\noindent\textbf{Dataset selection and model architectures.} We test our proposed techniques on CIFAR-10 (as the default dataset) and CIFAR-100 \cite{krizhevsky2009learning} datasets, utilizing PreResNet110 (as the default architecture), WideResNet-28-10, and Vision Transformer-base (ViT-base). 

\noindent\textbf{Evaluation methods and metrics.}  We set the attack strength parameters constraining the $\ell_\infty, \ell_2$, and $\ell_1$ norms to the commonly used values $\epsilon = 8/255, 1$, and $12$, respectively. In our evaluation, we implemented basic PGD adversarial attacks as well as Auto-Attack (AA) \cite{croce2020reliable} under $\ell_\infty$, $\ell_2$, $\ell_1$ norm perturbations, in addition to implementing the MSD attack. Metrics for assessment include: \ding{172} Standard accuracy (SA) on clean test data; \ding{173} Robust accuracies under various perturbation types including $\ell_\infty/\ell_2/\ell_1$-PGD, MSD attack, and $\ell_\infty/\ell_2/\ell_1$ AA; and \ding{174} Sample-wise worst-case scenario accuracy (Union) calculated from all three basic PGD adversarial methods. A sample is considered correct only if it is accurately predicted under each of the three basic PGD adversarial attacks. These experiments were run on two NVIDIA RTX A100 GPUs.

\begin{table}[ht]
\vspace*{-3mm}
\caption{\small Our Method Achieves State-Of-The-art Robustness Levels Under Various Perturbations on CIFAR-10. ERMC surpasses baseline performance without the use of an ensemble. The best results are in \textbf{bold}.}
\label{tab: main}
\vspace{-5mm}
\begin{center}
\resizebox{0.48\textwidth}{!}{
\begin{tabular}{l||c|c|c|c|c}
\hline
\hline
& SA & \begin{tabular}[c]{@{}c@{}} PGD \\ ($\ell_\infty$/$\ell_2$/$\ell_1$)  \end{tabular}  & Union & \begin{tabular}[c]{@{}c@{}} AA \cite{croce2020reliable} \\ ($\ell_\infty/\ell_2/\ell_1$)  \end{tabular}  & MSD \\
\hline
\begin{tabular}[c]{@{}c@{}} AT-$\ell_\infty$ \cite{madry2018towards}   \end{tabular}  &   85.00\% & 49.03\%/29.66\%/\{16.61\%\}  & 21.85\%  & 46.02\%/20.86\%/\{10.45\%\} & 15.27\% \\
\hline
\begin{tabular}[c]{@{}c@{}} MSD \cite{maini2020adversarial} \\ Defense  \end{tabular}  &  81.35\% & \{40.14\%\}/48.58\%/47.50\%  & 38.35\% & \{37.87\%\}/45.9\%/45.27\% & 38.20\%  \\
\hline
\begin{tabular}[c]{@{}c@{}} E-AT \cite{croce2022adversarial} \end{tabular}  &  79.3\% & \{44.07\%\}/49.12\%/49.82\% & 41.08\% & \{41.41\%\}/46.5\%/47.82\% & 42.67\%  \\
\hline
\begin{tabular}[c]{@{}c@{}} \textit{ERMC-1} \\ (ours $n=1$) \end{tabular}  &  82.66\%  &   \{46.54\%\}/48.76\%/47.06\% &  41.94\% & 44.88\%/45.88\%/\{43.97\%\} & 44.88\%   \\
\hline
\begin{tabular}[c]{@{}c@{}} \textit{ERMC-3} \\ (ours $n=3$)  \end{tabular}  &  79.61\%  & 49.29\%/51.32\%/\{48.49\%\} &  45.27\% & \{42.88\%\}/44.57\%/47.37\% & 43.31\%   \\
\hline
\begin{tabular}[c]{@{}c@{}} \textit{ERMC-5} \\ (ours $n=5$) \end{tabular}  &  79.41\%  & 55.46\%/57.28\%/{\{\bf{53.97\%}\}} &  \bf{51.41\%} & {\{\bf{49.33\%}\}}/50.55\%/52.41\% & {\bf{49.78\%}}   \\
\hline
\hline
\end{tabular}}
\end{center}
\end{table}

\vspace{-4mm}
\noindent\textbf{Experimental results.} As a baseline, endpoint models are trained for 150 epochs, with paths derived through an extra 50 epochs. The models at the left (right) endpoints are trained with AT-$\ell_\infty$ (AT-$\ell_\infty$ and fine-tuned with AT-$\ell_1$). The results are displayed in Fig.~\ref{fig: adv_self_rmc2}. The upper panels show the clean test accuracy and accuracies under $\ell_\infty/\ell_2/\ell_1$-PGD attacks. The lower panels show the corresponding loss values. $t$ varies from 0 to 1. Moving from left to right in Fig.~\ref{fig: adv_self_rmc2}, panels (a) and (b) depict results obtained from the CIFAR-10 and CIFAR-100 datasets, respectively, using the PreResNet110 model architecture. Conversely, panels (c) and (d) present results from the CIFAR-10 dataset, but utilizing the WideResNet-28-10 and ViT-base model architectures. Notably, we find: \ding{202} The existence of robust paths, which shows that the ERMC application enhances resilience to multiple attack types, although they don't form straight lines like in mode connectivity; \ding{203} ERMC performs well on all considered datasets and architectures; \ding{204} The robust paths also function as effective mode connectivity paths, where both the clean accuracy and loss (indicated by red lines) maintain consistent levels between the two endpoints $t=0$ and $t=1$; and \ding{205} Fine-tuning influences original robustness levels, where endpoint models show strong resilience against corresponding perturbation types but are weaker against others. For example, the left (right) endpoint has a high resilience to $\ell_\infty$ ($\ell_1$) perturbations but suffers from attacks using $\ell_1$ ($\ell_\infty$) perturbations. Additionally, ERMC reduces the required computation time by approximately $36\%$ on a single GPU relative to the brute force approach of solving AT-$\ell_\infty$ and AT-$\ell_1$ independently.


Comparative analyses with different baselines are summarized in Table~\ref{tab: main}. We evaluate them using all aforementioned metrics, and the lowest accuracy under the three basic $\ell_p$-PGD attacks (and three $\ell_p$ AA) are indicated within braces. These baselines - comprising AT-$\ell_\infty$ \cite{madry2018towards}, E-AT \cite{croce2022adversarial}, and MSD Defense \cite{maini2020adversarial} - are trained over 200 epochs. {The model selection thresholds are set at $\alpha_\infty=37\%$ for $\ell_\infty$ robustness and $\alpha_1=43\%$ for $\ell_1$ robustness.} As per Table~\ref{tab: main}, observe that ERMC-1 outperforms MSD Defense (and E-AT) in terms of accuracy improvements under various metrics, indicated by percentages $6.4\%$, $3.59\%$, $6.1\%$, and $6.68\%$ ($2.47\%$, $0.86\%$, $2.56\%$, and $2.21\%$) under $\ell_\infty/\ell_2/\ell_1$-PGD, Union, $\ell_\infty/\ell_2/\ell_1$ AA, and MSD Attack, respectively. It is also observable that as the number of models
$n$ increases, the performance of ERMC correspondingly improves. When $n$ reaches to 5, ERMC-5 outperforms MSD Defense (and E-AT) in terms of accuracy improvements under various metrics, indicated by percentages $13.85\%$, $13.06\%$, $11.46\%$, and $11.58\%$ ($9.9\%$, $10.33\%$, $7.92\%$, and $7.11\%$) under $\ell_\infty/\ell_2/\ell_1$-PGD, Union, $\ell_\infty/\ell_2/\ell_1$ AA, and MSD Attack, respectively. It's crucial to highlight that our method surpasses baseline performance without the use of an ensemble. The further enhancement observed with an ensemble simply underscores the value added by ensemble boosting of ERMC's baseline performance from the single model context. Unlike baselines that require multiple runs to generate a similar number of models, our approach naturally produces a model population in a single run, offering an attractive time-efficient alternative. The ERMC approach demonstrates a trade-off between clean accuracy and robustness. Nonetheless, the decrease in clean accuracy, quantified at $2.34\%$ when measured against AT-$\ell_\infty$, is more modest compared to the degradation suffered by other defensive strategies like MSD Defense and E-AT.

\section{Conclusion}\label{sec:conclusion}
This paper introduces the Efficient Robust Mode Connectivity (ERMC) method, a novel approach for enhancing the resilience of deep learning models against various adversarial $\ell_p$-norm attacks. By combining the robustness benefits of $\ell_1$ and $\ell_\infty$ adversarial training within a single framework, ERMC transcends the limitations of traditional methods that focus on single-type perturbations. Leveraging mode connectivity theory with efficient tuning and ensemble strategies, the method achieves a robust defense. Experimental results show that ERMC outperforms established defenses like AT-$\ell_\infty$, E-AT, and MSD Defense, particularly against $\ell_\infty$ and $\ell_1$ perturbations and other $\ell_p$-norm attacks. Its integration of multiple adversarial training types enhances defense capabilities while preserving efficiency, marking a significant step forward in adversarial robustness and suggesting new directions for further research in the security of deep learning.
\bibliographystyle{IEEEbib}
\bibliography{strings,refs,ref_adv}

\end{document}